\documentclass{article}

\usepackage[main, final]{neurips_2026}

\usepackage[utf8]{inputenc} 
\usepackage[T1]{fontenc}    
\usepackage{hyperref}       
\usepackage{url}            
\usepackage{booktabs}       
\usepackage{amsfonts}       
\usepackage{nicefrac}       
\usepackage{microtype}      
\usepackage[table]{xcolor}         
\usepackage{graphicx}
\usepackage{caption}
\usepackage{subcaption}
\usepackage{wrapfig}
\usepackage{multirow}
\usepackage{pifont}
\usepackage{amssymb}

\definecolor{easygreen}{RGB}{226,239,218}
\definecolor{hardred}{RGB}{248,215,218}
\newcommand{\cmark}{\ding{51}}
\newcommand{\xmark}{\ding{55}}

\title{Who Generated This 3D Asset? Learning Source Attribution for Generative 3D Models}

\author{
    Sihan Ma
    \quad\  {Siyuan Liang}
    \quad\  {Dacheng Tao}
    \vspace{0.5em}\\
    College of Computing \& Data Science, Nanyang Technological University, Singapore
    \vspace{0.5em}\\
    \texttt{\{sihan.ma, siyuan.liang, dacheng.tao\}@ntu.edu.sg}
}

\begin{document}
\maketitle

\begin{figure}[h]
  \centering
  \vspace{-10pt}
  \includegraphics[width=0.98\linewidth]{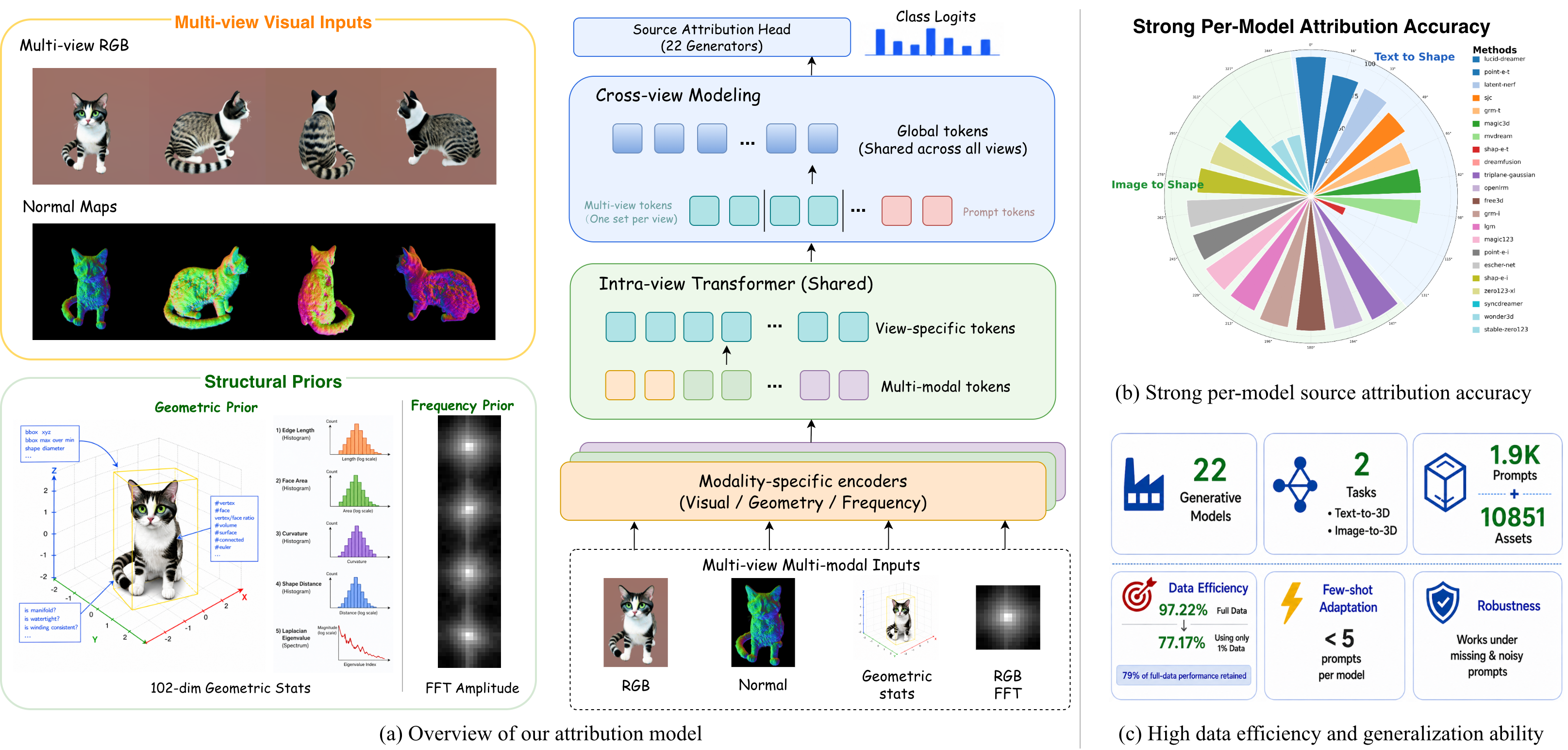}
  \caption{%
    \textbf{Overview of our attribution framework.}
    Given multi-view renderings and structural priors of a 3D asset, our model learns discriminative fingerprints through intra-view fusion and cross-view reasoning, enabling accurate attribution across 22 generative 3D models.
}
  \label{fig:teaser}
\vspace{-5pt}
\end{figure}

\begin{abstract}
Generative 3D models are deployed in gaming, robotics, and immersive creation, making source attribution critical: given a 3D asset, can we identify whether and which generative model created it?  This problem faces two core challenges: \textbf{dispersed attribution signals}, where 3D fingerprints are distributed across multi-view, geometric, and frequency-domain cues; and \textbf{realistic deployment constraints}, where scarce labels, degraded prompts, and mixed real/synthetic assets undermine attribution reliability.
To systematically study this problem, we construct, to the best of our knowledge, the first passive source attribution benchmark for modern generated assets, covering $22$ representative 3D generators under standard, few-shot, and realistic deployment protocols. Based on this benchmark, we find that generative 3D models leave two types of stable fingerprints: cross-view inconsistency and structural artifacts reflected in geometric statistics and frequency-domain cues. To capture these dispersed signals, we propose a hierarchical multi-view multi-modal Transformer that fuses appearance, geometric, and frequency-domain features within each view and models global relationships across views.
Extensive experiments demonstrate strong performance, achieving $97.22\%$ accuracy under full supervision and $77.17\%$ accuracy with only $1\%$ training data, corresponding to fewer than five samples per generator. 
These results show that modern 3D generators leave stable and attributable fingerprints, establishing a new benchmark and methodological foundation for trustworthy 3D content provenance.
\end{abstract}
\section{Introduction}
In recent years, generative 3D models have been rapidly transforming the way digital assets are produced. Text-to-3D and image-to-3D systems are now capable of generating diverse 3D objects at low cost and are gradually being integrated into real-world workflows such as game asset production~\cite{xu2025constraint,zang2025sketch2play}, robotics simulation~\cite{deitke,yang2024holodeck}, virtual world construction~\cite{engstler2025syncity,zhang2024text2nerf,hollein2023text2room}, and immersive content creation~\cite{dreamfusion,Magic3D,ProlificDreamer}.

At the same time, large-scale 3D asset libraries~\cite{Objaverse,Objaverse-XL,step1x} and open content ecosystems~\cite{nvidia_orca} are emerging, where synthetic assets can be downloaded, modified, reused, and redistributed. In robotics and embodied AI scenarios, synthetic 3D assets of unknown origin may also enter simulation environments and training datasets, thereby affecting data governance, quality auditing, and the reliability of downstream systems~\cite{zhu2025evading, janssen2020data, hassani2025mapping}. Therefore, 3D content provenance tracking is not merely a forensic problem but a foundational capability for deploying trustworthy 3D generative systems: \textbf{Given a 3D asset, can we determine whether it was generated by a generative model and further identify its specific source model?}

Although extensive research has studied attribution for 2D image and video generation~\cite{saga,wang2023did,deepfake}, \textbf{3D asset source attribution} still lacks a unified benchmark and systematic approach tailored to modern text-to-3D and image-to-3D generated assets. It faces two core challenges. 
First, \emph{dispersed attribution signals}. 3D assets are not single images; their source-model fingerprints may be distributed across multi-view appearances, cross-view relationships, geometric structures, and frequency-domain patterns. Single-view rendering or simple multi-view aggregation struggles to fully capture these clues. 
Second, \emph{realistic deployment constraints}. Attribution systems in real-world deployments often lack sufficient ground-truth data and may encounter missing or degraded prompts, as well as mixtures of real and synthetic assets, thereby undermining attribution reliability. 

To the best of our knowledge, we constructed the first passive source attribution benchmark for modern generative 3D assets. This benchmark is built on public resources such as 3DGen-Bench~\cite{3dgen-bench} and Cap3D~\cite{cap3d}, covering outputs from $22$ representative text-to-3D and image-to-3D generators. For evaluation, we design standard supervised, few-shot, and realistic deployment protocols to assess attribution capability, data efficiency, and robustness under prompt degradation and mixed real/synthetic scenarios.

Based on the benchmark, we find that modern generative 3D models leave behind two types of stable, complementary source-model fingerprints: \emph{cross-view inconsistency} (model-specific view artifacts that require explicit cross-view modeling) and complementary  \emph{structural artifacts} in the geometry and frequency domains. To capture these \emph{dispersed attribution signals}, we propose a hierarchical multi-view, multi-modal Transformer. The model first fuses RGB appearance, geometric descriptors, and frequency-domain features within each view to capture local appearance and structural fingerprints; it then models global relationships across views to identify generator-specific cross-view inconsistencies. For \emph{realistic deployment constraints}, rather than presenting them as an additional methodological component, we evaluate the reliability of the method under constrained deployment conditions through few-shot learning, prompt degradation, and mixed real/synthetic protocols.

Extensive experiments demonstrate that our method performs consistently on source attribution across $22$ generators. Under the standard fully supervised setting, our method achieves an attribution accuracy of $97.22\%$; under extremely low-label conditions, using only $1\%$ of the training data, i.e., fewer than $5$ samples per generator, it still achieves an accuracy of $77.17\%$. Further experiments show that the method remains robust when prompts are missing or noisy and when real and synthetic assets are mixed. These results indicate that modern generative 3D models indeed leave source-model fingerprints that can be reliably captured, providing a new benchmark and methodological foundation for the attribution, auditing, and governance of trustworthy 3D-generated content. Our contributions are summarized as follows:
\begin{itemize}
    \item We introduce passive source attribution for modern generative 3D assets and present the first unified benchmark (to our knowledge) spanning 22 text-to-3D and image-to-3D generators across standard, few-shot, and realistic deployment protocols.
    \item We identify two stable fingerprints--cross-view inconsistencies and geometric/frequency structural artifacts--and propose a hierarchical multi-view multi-modal Transformer to capture these dispersed signals.
    \item Our method achieves $97.22\%$ accuracy across $22$ generators under full supervision and $77.17\%$ accuracy with only $1\%$ training data, while remaining robust to degraded prompts and mixed real/synthetic settings.

\end{itemize}

\section{Related Work}

\subsection{Generative 3D Models and Benchmarks}

Recent advances in diffusion models, neural rendering, and large-scale data have driven rapid progress in generative 3D modeling. DreamFusion~\cite{dreamfusion} first showed that pretrained 2D diffusion models can be adapted for text-to-3D synthesis via score distillation. Subsequent works such as Magic3D~\cite{Magic3D} improved quality and efficiency through coarse-to-fine optimization, while feed-forward approaches like Point-E~\cite{Ponit-e} and Shap-E~\cite{Shap-e} enabled faster generation using point-based or implicit representations. These methods span diverse paradigms (optimization-based, feed-forward, reconstruction-driven) and support different input modalities, including text and images.

In parallel, large-scale 3D data resources have expanded significantly. Datasets such as Objaverse~\cite{Objaverse}, Objaverse-XL~\cite{Objaverse-XL} provide millions of assets, while text-aligned datasets like Cap3D~\cite{cap3d} support multimodal learning. 3DGen-Bench~\cite{3dgen-bench} provides standardized evaluation protocols for both text-to-3D and image-to-3D generation, enabling comparison across models under diverse prompts and categories. These resources have enabled benchmarks focusing on generation quality, diversity, and controllability. However, prior efforts largely evaluate fidelity and reconstruction performance, while the provenance and source attribution of generated 3D assets remain underexplored.

\subsection{AI Content Provenance and Source Attribution}

\noindent\textbf{AI-generated media provenance and accountability.}
The proliferation of generative models has spurred research on authenticity analysis and provenance tracing~\cite{wang2020cnn, 
rossler2019faceforensics, corvi2023detection,liang2024poisoned}. Prior work shows that generative models leave identifiable traces, enabling forensic analysis and attribution~\cite{marra2019gans, 
yu2019attributing, yu2021artificial, girish2021towards}. For example, GAN-generated images exhibit model-specific fingerprints~\cite{yu2019attributing}, while later studies explored training-data-based attribution and scalable fingerprinting for responsible model release~\cite{yu2021artificial,liu2022responsible}. These methods highlight the importance of provenance for governance and accountability, but mainly focus on 2D media and often address binary real-vs-fake detection or model tracing in rasterized outputs.

\noindent\textbf{Fine-grained source attribution and the gap to 3D assets.}
Beyond binary detection, recent work studies fine-grained source attribution, aiming to identify the exact generator among multiple candidates~\cite{yu2019attributing, 
yu2021artificial, girish2021towards,albright2019source}. Existing approaches address both closed-set and open-world settings for GAN-generated images~\cite{yu2019attributing, girish2021towards}, as well as open-set attribution under unseen generators~\cite{yang2023progressive}. More recent studies analyze how generative models encode source-specific fingerprints~\cite{song2024manifpt}. 
In the 3D domain, the closest existing attribution work, FAKEPCD~\cite{fakepcd}, mainly focuses on authenticity detection and source attribution for synthetic point clouds~\cite{fakepcd}, while watermarking and content credential methods~\cite{chen2023universal} typically rely on actively embedding source signals during generation or preserving trusted metadata. In contrast, this paper investigates a more open \emph{passive source attribution} setting: identifying the origin of a 3D asset already in circulation, without assuming 
watermarks~\cite{chen2023universal} or active copyright-side defenses
~\cite{guo2024copyrightshield, yang2025me}.

\section{Attribution Model}

\noindent\textbf{Problem setup.}
We study 3D asset source attribution: given a released synthetic 3D asset and optional metadata, the goal is to infer which generative model produced it. We primarily consider attribution over a candidate pool of 22 representative 3D generators spanning both text-to-3D and image-to-3D paradigms, diverse architectures, and multiple representation families. Beyond this primary closed-set setting, we additionally consider realistic deployments where metadata may be incomplete and synthetic assets may coexist with real scanned objects.

\noindent\textbf{Motivation and risk scenario.}
As AI-generated 3D assets are increasingly reused and redistributed, information about their origins can easily be lost or hidden. For example, assets may be uploaded without disclosure, claimed as manually created~\cite{yu2021artificial,liu2022responsible,liang2022imitated}, or incorporated into datasets without clear provenance~\cite{longpre2024large,shumailov2024collapse}.

When source information is unavailable, platforms and users may still need to determine an asset's origin for auditing, governance, and accountability. 
This motivates \emph{passive attribution} directly from the released asset itself, without assuming access to generation logs or trusted provenance metadata.

\noindent\textbf{Defender's observations and assumptions.}
The defender is given the released 3D asset and may optionally access associated metadata when available, but does not rely on metadata being complete or trustworthy. 
The defender does not have access to the original prompt, generation logs, training data, or the source generator's internal parameters. 
Instead, the defender analyzes observable evidence derived from the asset, including rendered multi-view appearance, geometric statistics, frequency-domain patterns, and cross-view relationships. 

Although any single cue may be weak or incomplete, generative models can leave recurring signatures due to differences in architectures, training data, representations, and generation pipelines~\cite{marra2019gans,corvi2023intriguing,frank2020leveraging,durall2020watch,asnani2023reverse}. 
These complementary signals provide the basis for post-hoc attribution from the released asset alone.

\noindent\textbf{Challenges in 3D asset attribution.}
Reliable attribution remains challenging due to both \emph{dispersed attribution signals} and \emph{realistic deployment constraints}. 
Unlike 2D images, attribution cues in generated 3D assets are often distributed across viewpoints, renderings, and structural representations, making reliable attribution difficult from isolated observations alone. 

At the same time, practical deployment may involve incomplete or corrupted metadata, scarce supervision, and mixtures of real and synthetic assets. 
These challenges motivate attribution models that can jointly reason across views and modalities while remaining robust under degraded real-world conditions.
\section{Benchmark}\label{sec:benchmark}

\begin{figure*}[t]
\centering
\vspace{-10pt}
\begin{minipage}[t]{0.45\textwidth}
\vspace{0pt}
\centering
\scriptsize

\captionof{table}{Evaluation protocols.}
\label{tab:benchmark_protocols}

\setlength{\tabcolsep}{4pt}

\begin{tabular}{lcccc}
\toprule
Protocol & Prompt & Real & \#Data & Goal \\
\midrule
Standard       & Clean     & \xmark & Full     & Base \\
Few-shot       & Clean     & \xmark & 1\%  & Efficiency \\
Missing Prompt & Empty     & \xmark & 1\%     & Prompt-free \\
Noisy Prompt   & Corrupted & \xmark & 1\%     & Robustness \\
Real-Synthetic     & Clean     & \cmark & 1\%     & Deployment \\
\bottomrule
\end{tabular}

\vspace{0.3em}

\end{minipage}
\hfill
\begin{minipage}[t]{0.5\textwidth}
\vspace{0pt}
\centering

\includegraphics[width=0.85\linewidth]{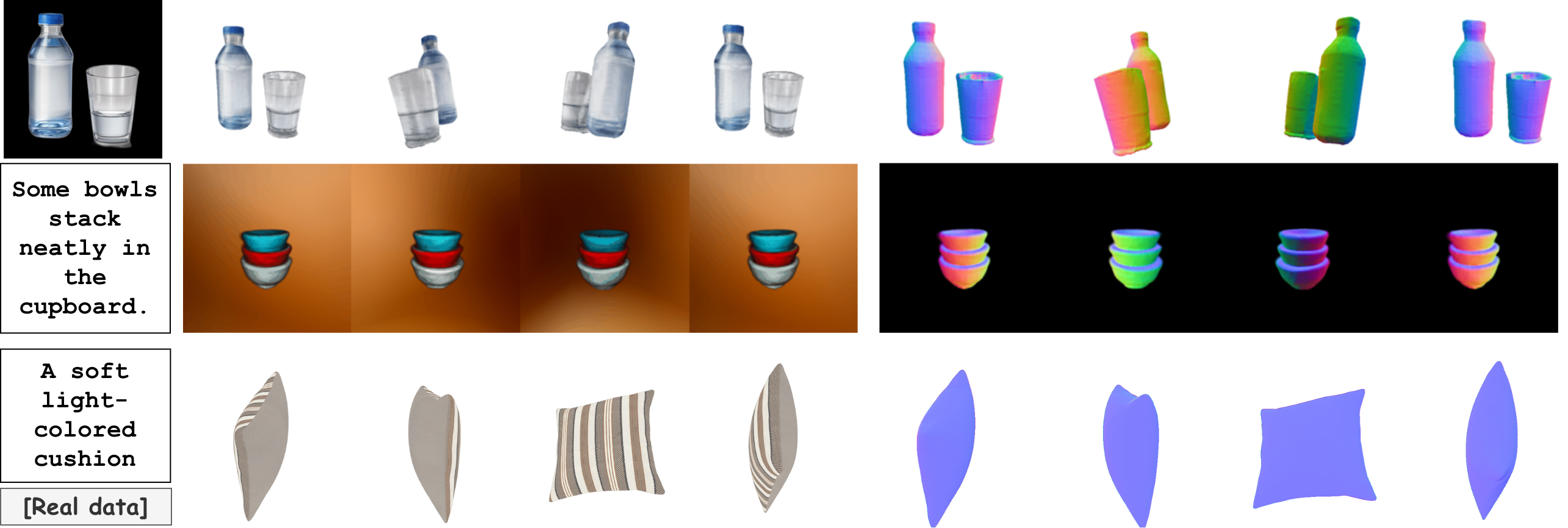}

\captionof{figure}{
Representative benchmark samples.
}
\label{fig:benchmark_samples}

\end{minipage}
\vspace{-10pt}
\end{figure*}

We introduce the first benchmark for passive 3D asset source attribution, spanning diverse generators, unified 3D representations, and realistic evaluation protocols. 
Tab.~\ref{tab:benchmark_protocols} summarizes the evaluation settings. Fig.~\ref{fig:benchmark_samples} demonstrates representative visual examples. See Sec.~\ref{sec:appendix-benchmark} for more details.

\noindent\textbf{Benchmark construction.}
Our benchmark is built upon public resources including 3DGen-Bench~\cite{3dgen-bench} and Cap3D~\cite{cap3d}. 
The synthetic subset covers 22 representative 3D generators spanning both text-to-3D and image-to-3D paradigms, with diverse architectures and underlying 3D representations. 
Overall, the benchmark contains 1,900 prompts and 10,851 synthetic assets. 
To enable unified attribution analysis across heterogeneous outputs, all assets are standardized into a common PLY representation and rendered into consistent multi-view observations. 
We further augment each asset with attribution-oriented structural priors, including normal maps, geometric descriptors, and Fast Fourier Transform (FFT)-based frequency cues~\cite{corvi2023intriguing,frank2020leveraging}. 

Beyond synthetic assets, we additionally construct a real-world subset using scanned 3D data from Cap3D~\cite{cap3d}, enabling evaluation under mixed real/synthetic deployment scenarios. 
To improve robustness evaluation, we further introduce prompt degradation protocols including prompt sparsification, masking, and corruption, simulating incomplete or noisy metadata conditions frequently encountered in practical applications.

\noindent\textbf{Evaluation protocols.}
We evaluate source attribution under multiple deployment-oriented protocols, including standard full-supervision, few-shot learning, missing-prompt, noisy-prompt, and mixed real/synthetic settings. 
Unless otherwise specified, experiments are primarily conducted under the 1\% training-data regime, simulating scarce attribution supervision in realistic scenarios. 
The standard protocol evaluates attribution under clean prompts and full training data, while the few-shot protocol focuses on label efficiency under limited supervision. 
The missing-prompt and noisy-prompt settings evaluate robustness against incomplete or corrupted metadata, respectively. 
Finally, the real-synthetic protocol evaluates attribution performance when generated assets are mixed with real scanned data.
\section{Method}\label{sec:method}

\subsection{Problem Formulation}

Given a released 3D asset $A$, we consider a passive attribution setting where the goal is to predict its source identity $y$ from observable cues $\mathcal{O}(A)$ and optional metadata $m$ (e.g., text prompts or reference images), i.e., $f_{\theta}(\mathcal{O}(A), m)\rightarrow y$.

Rather than directly operating on raw 3D representations, we formulate attribution over observable cues for two reasons: (i) generated assets may adopt heterogeneous formats (e.g., meshes, implicit fields, Gaussian-based representations), making direct comparison difficult; and (ii) attribution-relevant signals are often dispersed and reflected in rendered observations and derived structural cues, enabling a unified and deployment-friendly formulation.

In the default setting, $\mathcal{O}(A)$ consists of multi-view RGB renderings $R=\{r_i\}_{i=1}^{V}$ and corresponding normal renderings $N=\{n_i\}_{i=1}^{V}$, where $V=4$ unless otherwise specified. The default label space is $\mathcal{Y}_{\mathrm{syn}}=\{g_1,\dots,g_{22},u\}$, where $g_i$ denotes a known generator and $u$ denotes an unknown-generator class. In mixed real-and-synthetic settings, we further extend the label space as $\mathcal{Y}_{\mathrm{mix}}=\{g_1,\dots,g_{22},u,r\}$, where $r$ denotes real assets.
This formulation naturally supports missing metadata, unknown generators, and mixed real-synthetic deployment scenarios.

\begin{figure}[t]
  \centering
  \includegraphics[width=\linewidth]{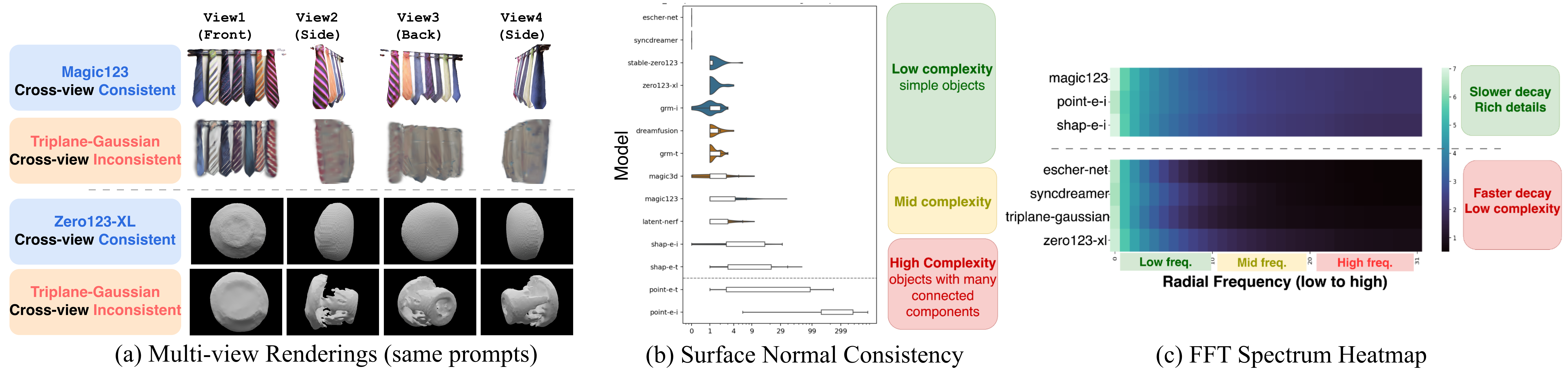}
  \caption{%
    Dispersed attribution fingerprints in generated 3D assets.
}
  \label{fig:method-challenges}
\end{figure}

\subsection{Attribution Fingerprints in Generative 3D Assets}\label{sec:method-fingerprints}

Despite the dispersed nature of attribution signals, we observe that modern generative 3D models leave behind stable and complementary source-model fingerprints. 

\noindent\textbf{Cross-view inconsistency patterns.} As shown in Fig.~\ref{fig:method-challenges}(a), different models often produce characteristic behaviors across viewpoints, such as hidden-surface collapse, view-dependent structural inconsistency, or stable multi-view geometry. These signals are difficult to identify from isolated renderings alone, but become apparent when modeling relationships across views.

\noindent\textbf{Structural artifacts.}
Generators exhibit complementary \emph{structural artifacts} beyond appearance. In Fig.~\ref{fig:method-challenges}(b), different generators produce distinct geometric regularization patterns, reflected in surface smoothness and structural consistency statistics. In Fig.~\ref{fig:method-challenges}(c), generators also show characteristic frequency-domain signatures, including different spectral decay behaviors and high-frequency distributions. Unlike view-specific appearance cues, these structural signals are often more globally consistent across viewpoints.

Together, these observations suggest that reliable attribution requires jointly modeling appearance, structural, and relational signals across multiple views and modalities, motivating a hierarchical multi-view multi-modal attribution framework.

\begin{figure}[t]
  \centering
  \includegraphics[width=\linewidth]{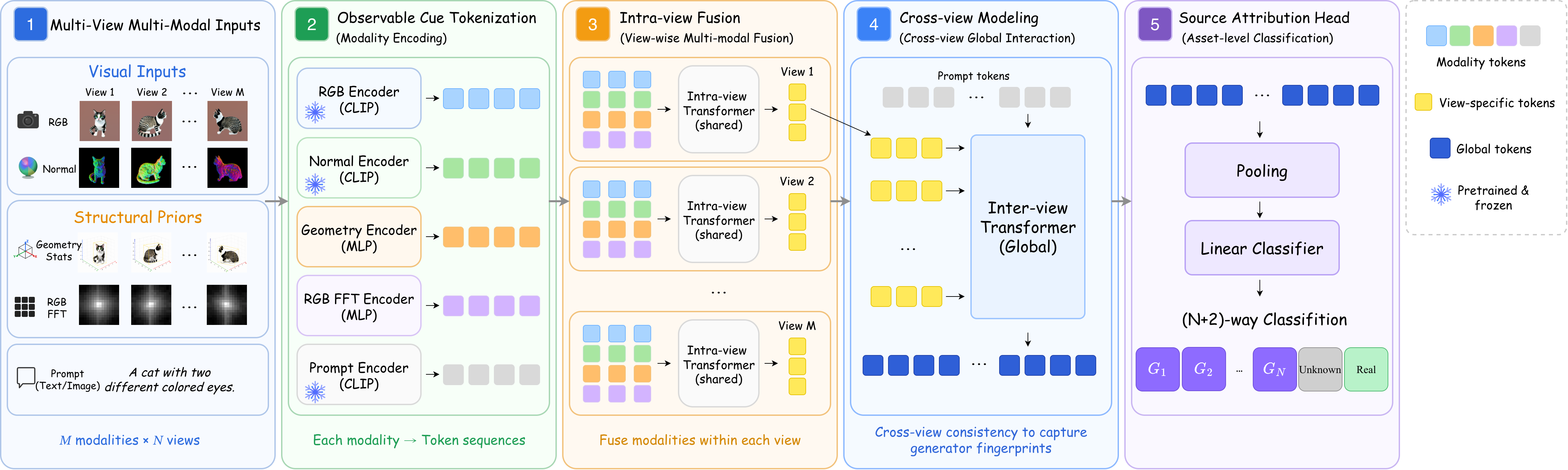}
  \caption{%
    \textbf{Overview of our attribution model.}
    Given multi-view renderings and structural priors from a 3D asset, our model learns dispersed attribution signals across viewpoints and modalities through multi-modal fusion within each single view and cross-view reasoning, achieving strong attribution accuracy and interpretable clustering across 22 generative 3D models.
}
  \label{fig:arch}
\end{figure}

\subsection{Capturing Dispersed Signals with Hierarchical Multi-view Modeling}

Motivated by the observations in Sec.~\ref{sec:method-fingerprints}, we propose a hierarchical multi-view multi-modal Transformer for modeling dispersed attribution signals across viewpoints and modalities. As shown in Fig.~\ref{fig:arch}, the framework jointly captures view-specific appearance cues, complementary structural artifacts, and generator-specific cross-view inconsistencies.

\noindent\textbf{Observable cue tokenization.}
For each viewpoint $i$, we construct a view-conditioned observation $x_i=\{r_i,n_i,s_i,q_i\}$, where $r_i$ and $n_i$ denote RGB and normal renderings, $s_i$ denotes geometry-derived structural descriptors, and $q_i$ denotes FFT-based frequency features. Optional metadata is denoted by $m$. Each modality is encoded into token embeddings via modality-specific encoders, yielding per-view multi-modal tokens together with optional metadata tokens. Specifically, RGB, normal, and metadata modalities are encoded using pretrained frozen vision-language encoders, while structural and frequency cues are projected using lightweight learnable MLPs.

\noindent\textbf{Intra-view multi-modal fusion.}
Within each viewpoint, modality tokens are fused using a shared Transformer to capture complementary local evidence across appearance, geometry, and frequency domains.

\noindent\textbf{Cross-view relationship modeling.}
The fused view representations are further aggregated by a global Transformer to model relationships across viewpoints, enabling the identification of generator-specific cross-view inconsistencies and complementary structural patterns that cannot be inferred from isolated views alone.

\noindent\textbf{Source attribution head.}
The aggregated representation is fed into a linear classification head for source prediction. The framework is trained end-to-end using cross-entropy loss. During training, metadata inputs may be randomly dropped, improving robustness to missing prompts at test time. Implementation details are provided in the Appendix Sec.~\ref{sec:appendix-method} and Sec.~\ref{sec:appendix-experiment}.

\subsection{Robust Attribution under Realistic Deployment}\label{sec:method-robust}

To support realistic deployment beyond standard attribution, we consider the following settings.

\noindent\textbf{Closed-set attribution with unknown generators.}
We adopt a standard attribution setting over 22 known generators, while additionally introducing an unknown-generator class to account for unseen or out-of-distribution sources encountered after deployment.

\noindent\textbf{Few-shot attribution.}
Real-world attribution may lack sufficient labeled training data, especially for newly emerging generators. To evaluate robustness under limited supervision, we consider few-shot attribution with only 1\% labeled training data.

\noindent\textbf{Metadata-optional attribution.}
Practical deployments cannot assume reliable prompts or reference metadata. We therefore treat metadata as optional inputs and evaluate attribution under sparse, occluded, or missing prompts through metadata corruption and masking. During training, metadata inputs may be randomly dropped when available, improving robustness to degraded metadata.

\noindent\textbf{Mixed real-and-synthetic attribution.}
Open-world environments may contain both generated and real assets. We therefore extend the label space with a real-asset category and evaluate attribution in mixed real-and-synthetic settings.

\section{Experiment}
\subsection{Experiment Setup}\label{sec:experiment-setup}

\noindent\textbf{Benchmark and protocols.}
We evaluate all methods on the benchmark introduced in Sec.~\ref{sec:benchmark}, under the realistic deployment protocols described in Sec.~\ref{sec:method-robust}.

\noindent\textbf{Baselines.}
Since no standardized baseline exists for 3D asset attribution, we construct generic attribution baselines under a unified multi-view formulation. 
We compare against \textsc{Grid-MLP}, \textsc{Grid-CNN}, and \textsc{Grid-Trans}, which process grid-based multi-view RGB renderings, normal renderings, and metadata using standard MLP, CNN, and Transformer backbones, respectively. 
Unlike our hierarchical formulation, these baselines aggregate multi-view evidence only implicitly and do not explicitly model cross-view or structural relationships.

\noindent\textbf{Metrics and implementation details.}
We report accuracy (Acc.), precision (Prec.), recall (Rec.), and F1-score. All models are trained end-to-end using AdamW~\cite{adamw}. Additional implementation details are provided in the Sec.~\ref{sec:appendix-experiment} in the Appendix.

\begin{table*}[t]
\centering
\caption{Comparison under few-shot and full-data settings. All values are reported in percentage (\%). ``\colorbox{black!5}{Text}'' and ``\colorbox{black!5}{Image}'' denote attribution accuracy on \textbf{text}-to-3D and \textbf{image}-to-3D generators.}
\scriptsize
\label{tab:model_compare}
\setlength{\tabcolsep}{4pt}
\renewcommand{\arraystretch}{1.15}
\resizebox{0.9\textwidth}{!}{
\begin{tabular}{l|cccccc|cccccc}
\toprule
\multirow{2}{*}{Model} &
\multicolumn{6}{c|}{1\% Data} &
\multicolumn{6}{c}{Full Data} \\
\cmidrule(lr){2-7} \cmidrule(lr){8-13}
&
Acc. & Prec. & Rec. & F1 & \cellcolor{black!5}{Text} & \cellcolor{black!5}{Image} &
Acc. & Prec. & Rec. & F1 & \cellcolor{black!5}{Text} & \cellcolor{black!5}{Image} \\
\midrule

\textsc{Grid-CNN}
& 30.76 & 27.04 & 30.97 & 26.70 & 36.60 & 27.12
& 35.34 & 35.20 & 35.13 & 27.84 & 46.84 & 28.15 \\

\textsc{Grid-MLP}
& 47.10 & 46.05 & 47.16 & 43.22 & 47.29 & 46.99
& 51.68 & 55.36 & 51.83 & 49.04 & 62.05 & 45.20 \\

\textsc{Grid-Trans.}
& 54.52 & 56.60 & 54.42 & 52.74 & 50.90 & 56.78
& 92.93 & 93.30 & 92.97 & 93.00 & 94.88 & 91.71 \\

Ours
& \textbf{77.17} & \textbf{79.08} & \textbf{76.98} & \textbf{74.78} & \textbf{68.67} & \textbf{82.49}
& \textbf{97.22} & \textbf{97.36} & \textbf{97.26} & \textbf{97.25} & \textbf{97.14} & \textbf{97.27} \\

\bottomrule
\end{tabular}
}
\end{table*}
\begin{table}[t]
  \centering
  \vspace{-10pt}
  \footnotesize
  \setlength{\tabcolsep}{8pt}
  \caption{Per-source performance analysis across different generative models. \colorbox{easygreen}{Green} and \colorbox{hardred}{red} cells indicate highly reliable and challenging attribution cases, respectively. See full table at Tab.~\ref{tab:appendix-per-source}}
  \label{tab:per-source-acc-full}
  \begin{tabular}{ll cccc cccc}
    \toprule
    & & \multicolumn{4}{c}{1\% Data} & \multicolumn{4}{c}{Full Data} \\
    \cmidrule(r){3-6} \cmidrule(l){7-10}
    Selected Model & Task & Prec. & Rec. & F1 & Acc. & Prec. & Rec. & F1 & Acc. \\
    \midrule
    L-Dreamer~\cite{Luciddreamer} & Text & \cellcolor{green!12}{100.0} & \cellcolor{green!12}{97.47} & \cellcolor{green!12}{98.72} & \cellcolor{green!12}{100.0} & \cellcolor{green!12}{100.0} & \cellcolor{green!12}{100.0} & \cellcolor{green!12}{100.0} & \cellcolor{green!12}{100.0} \\
    GRM-T~\cite{GRM} & Text & 75.34 & \cellcolor{green!12}{94.83} & 83.97 & 75.34 & \cellcolor{green!12}{100.0} & \cellcolor{green!12}{100.0} & \cellcolor{green!12}{100.0} & \cellcolor{green!12}{100.0} \\
    MVDream~\cite{Mvdream} & Text & 78.67 & \cellcolor{green!12}{93.65} & 85.51 & 78.67 & \cellcolor{green!12}{98.59} & \cellcolor{green!12}{93.33} & \cellcolor{green!12}{95.89} & \cellcolor{green!12}{93.33} \\
    Shap-E-T~\cite{Shap-e} & Text & \cellcolor{red!10}{26.32} & 86.96 & 40.40 & \cellcolor{red!10}{26.32} & 89.41 & \cellcolor{green!12}{100.0} & \cellcolor{green!12}{94.41} & \cellcolor{green!12}{100.0} \\
    DreamFusion~\cite{dreamfusion} & Text & \cellcolor{red!10}{6.41} & 83.33 & \cellcolor{red!10}{11.90} & \cellcolor{red!10}{6.41} & \cellcolor{green!12}{97.40} & \cellcolor{green!12}{96.15} & \cellcolor{green!12}{96.77} & \cellcolor{green!12}{96.15} \\
    \midrule
    Triplane-Gau.~\cite{Triplane} & Image & \cellcolor{green!12}{98.77} & \cellcolor{green!12}{94.12} & \cellcolor{green!12}{96.39} & \cellcolor{green!12}{98.77} & \cellcolor{green!12}{100.0} & \cellcolor{green!12}{100.0} & \cellcolor{green!12}{100.0} & \cellcolor{green!12}{100.0} \\
    OpenLRM~\cite{OpenLRM} & Image & \cellcolor{green!12}{96.43} & 79.41 & 87.10 & \cellcolor{green!12}{96.43} & \cellcolor{green!12}{100.0} & \cellcolor{green!12}{100.0} & \cellcolor{green!12}{100.0} & \cellcolor{green!12}{100.0} \\
    Magic123~\cite{Magic123} & Image & \cellcolor{green!12}{91.46} & 64.10 & 75.38 & \cellcolor{green!12}{91.46} & \cellcolor{green!12}{98.80} & \cellcolor{green!12}{100.0} & \cellcolor{green!12}{99.39} & \cellcolor{green!12}{100.0} \\
    Zero123-XL~\cite{Zero-1-to-3} & Image & 76.25 & 64.21 & 69.71 & 76.25 & 89.41 & \cellcolor{green!12}{95.00} & \cellcolor{green!12}{92.12} & \cellcolor{green!12}{95.00} \\
    Wonder3D~\cite{Wonder3D} & Image & {45.68} & 84.09 & 59.20 & {45.68} & \cellcolor{green!12}{96.30} & \cellcolor{green!12}{96.30} & \cellcolor{green!12}{96.30} & \cellcolor{green!12}{96.30} \\
    \midrule
    Average (all 22) & -- & 79.08 & 76.98 & 71.53 & 77.17 & \cellcolor{green!12}{93.13} & \cellcolor{green!12}{93.03} & \cellcolor{green!12}{93.02} & \cellcolor{green!12}{97.22} \\
    \bottomrule
  \end{tabular}
  \vspace{-6pt}
\end{table}
\subsection{Main Results}

\noindent\textbf{Overall comparison.}
Tab.~\ref{tab:model_compare} shows that our hierarchical multi-view formulation substantially outperforms all generic attribution baselines under both few-shot and standard full-data settings. Under the challenging 1\% training regime, our model achieves 77.17\% accuracy, outperforming \textsc{Grid-Trans} by +22.65\% and \textsc{Grid-CNN} by +46.41\%, demonstrating strong data efficiency under limited supervision. Under full-data training, our model further reaches 97.22\% accuracy and consistently improves all metrics over Transformer baselines, indicating that explicitly modeling cross-view relationships and structural cues is substantially more effective than implicit multi-view aggregation.

\noindent\textbf{Per-source analysis.}
Tab.~\ref{tab:per-source-acc-full} shows that our model achieves strong and consistent attribution performance across diverse generator families. Under the 1\% setting, the model remains highly robust on most image-to-shape generators, with many sources exceeding 90\% accuracy despite extremely limited supervision. More importantly, our model substantially improves attribution for structurally challenging generators such as DreamFusion~\cite{dreamfusion} and Shap-E~\cite{Shap-e}, which are difficult for generic appearance-based baselines due to severe cross-view inconsistency and geometry artifacts. The performance gap is particularly pronounced in few-shot setting, suggesting that the proposed hierarchical modeling captures stable generator-specific fingerprints beyond superficial appearance cues.

\begin{table}[t]
\vspace{-10pt}
  \centering
  \scriptsize
  \setlength{\tabcolsep}{4pt} 
  \caption{Ablation study results. \textbf{Rendering}, \textbf{Geometry}, \textbf{MVC}, and \textbf{Hierarchical} denote Rendering (RGB \& Normal), Geometry \& frequency, Multi-view Consistency, Hierarchical fusion, respectively. ``\colorbox{black!5}{Text}'' and ``\colorbox{black!5}{Image}'' denote attribution accuracy on \textbf{text}-to-3D and \textbf{image}-to-3D generators.}
  \label{tab:ablation-study}
    \begin{tabular}{l cccc cccccc}
    \toprule
    ID & Rendering & Geometry & MVC & Hierarchical & Acc. & Prec. & Rec. & F1 & \cellcolor{black!5}{Text} & \cellcolor{black!5}{Image} \\
    \midrule
    1 & \checkmark & & & & 54.52 &	56.6	& 54.42 &	52.74 &	50.9 &	56.78\\
    1 & \checkmark & \checkmark &        &        & 64.66 & 64.20 & 64.43 & 62.43 & 59.64 & 67.80 \\
    2 & \checkmark &            & \checkmark &        & 69.47 & 70.88 & 69.55 & 67.31 & 61.45 & 74.48 \\
    3 & \checkmark & \checkmark & \checkmark &        & 75.26 & 75.64 & 75.17 & 73.27 & 65.96 & 81.07 \\
    4 & \checkmark & \checkmark & \checkmark & \checkmark & \textbf{77.17} & \textbf{79.08} & \textbf{76.98} & \textbf{74.78} & \textbf{68.67} & \textbf{82.49} \\
    \bottomrule
    \end{tabular}
    \vspace{-4pt}
\end{table}

\subsection{Ablation Study}
Tab.~\ref{tab:ablation-study} demonstrates that each proposed component contributes substantially to attribution performance. Starting from the rendering-only baseline (\textsc{Grid-Trans}, 54.52\% Acc.), adding geometry and frequency priors improves accuracy by +10.14\%, confirming that structural cues provide strong complementary fingerprints beyond appearance alone. Replacing implicit grid concatenation with explicit multi-view consistency modeling further boosts performance to 69.47\%, showing the importance of separately tokenizing views to preserve cross-view relational evidence. Combining structural priors with explicit multi-view modeling yields a large gain to 75.26\%, while the full hierarchical formulation further reaches 77.17\% accuracy by jointly capturing view-wise modalities and global cross-view interactions. Additional ablations on the number of views are provided in the Appendix.

\begin{table*}[t]
\centering
\scriptsize
\caption{Robust attribution under realistic deployment settings. ``*'' indicates removing prompts during both training and inference. See Tab.~\ref{tab:appendix-prompt-dependency} for more results.}
\label{tab:deployment_robustness}
\begin{tabular}{clcccc}
\toprule
Task / Protocol & Setting & Prec. & Rec. & F1 & Acc. \\
\midrule

\multirow{4}{*}{Sparse Text Prompt}
& Full        & 85.06 & \textbf{69.17} & 69.97 & \textbf{68.67} \\
& Sparse (4 words) & 85.35 & 68.47 & \textbf{70.25} & 67.92 \\
& Sparse (1 word) & 85.09 & 64.16 & 67.27 & 63.55 \\
& Empty       & \textbf{87.29} & 63.74 & 67.08 & 63.25 \\
& Empty*      & 84.14 & 61.50 & 66.24 & 60.69 \\

\midrule

\multirow{4}{*}{
\begin{tabular}{c}
Noisy Image Prompt \\
{\scriptsize (Gaussian noise std. $\sigma$)}
\end{tabular}
}
& Clean       & 84.15 & \textbf{82.39} & 82.16 & \textbf{82.49} \\
& $\sigma=32$ & 85.15 & 82.38 & 82.96 & \textbf{82.49} \\
& $\sigma=96$ & \textbf{85.53} & 81.81 & \textbf{82.71} & 81.92 \\
& Empty*      & 84.27 & 78.62 & 80.10 & 78.72 \\

\midrule

\multirow{3}{*}{
\begin{tabular}{c}
Masked Image Prompt \\
{\scriptsize (Masked ratio $r$)}
\end{tabular}
}
& $r=20\%$ & 50.43 & \textbf{48.75} & \textbf{49.02} & \textbf{82.58} \\
& $r=60\%$ & 50.01 & 45.91 & 47.29 & 77.78 \\
& $r=90\%$ & \textbf{50.72} & 43.69 & 46.19 & 74.01 \\

\midrule

\multirow{2}{*}{Real-Synthetic}
& Synthetic only & \textbf{79.08} & 76.98 & 71.53 & 77.17 \\
& w/ Real        & \textbf{79.08} & \textbf{77.54} & \textbf{75.68} & \textbf{78.90} \\

\bottomrule
\end{tabular}
\vspace{-12pt}
\end{table*}

\subsection{Robust Attribution under Realistic Deployment Settings}
Tab.~\ref{tab:deployment_robustness} shows that our model remains highly robust under degraded metadata and mixed real-world conditions. Under sparse text prompts, attribution performance degrades gracefully from 68.67\% to 60.69\% accuracy even when prompts are entirely removed, indicating that the model does not overly rely on metadata shortcuts and can instead leverage stable structural and cross-view fingerprints. 

For image-to-shape attribution, the model is remarkably insensitive to prompt corruption. Even under severe image noise ($\sigma=96$), performance remains nearly unchanged (82.49\% $\rightarrow$ 81.92\% Acc.), suggesting that attribution signals are dominated by generator-specific structural patterns rather than prompt fidelity. Similarly, under aggressive image masking ($r=90\%$), the model still achieves 74.01\% accuracy, demonstrating strong robustness to partial or heavily degraded visual conditions.

Finally, introducing real assets further improves overall attribution performance from 77.17\% to 78.90\%, indicating that the proposed formulation generalizes effectively beyond purely synthetic closed-set environments and remains reliable in mixed real-and-synthetic deployment scenarios.

\begin{figure}[t]
  \vspace{-10pt}
  \centering
  \includegraphics[width=0.95\linewidth]{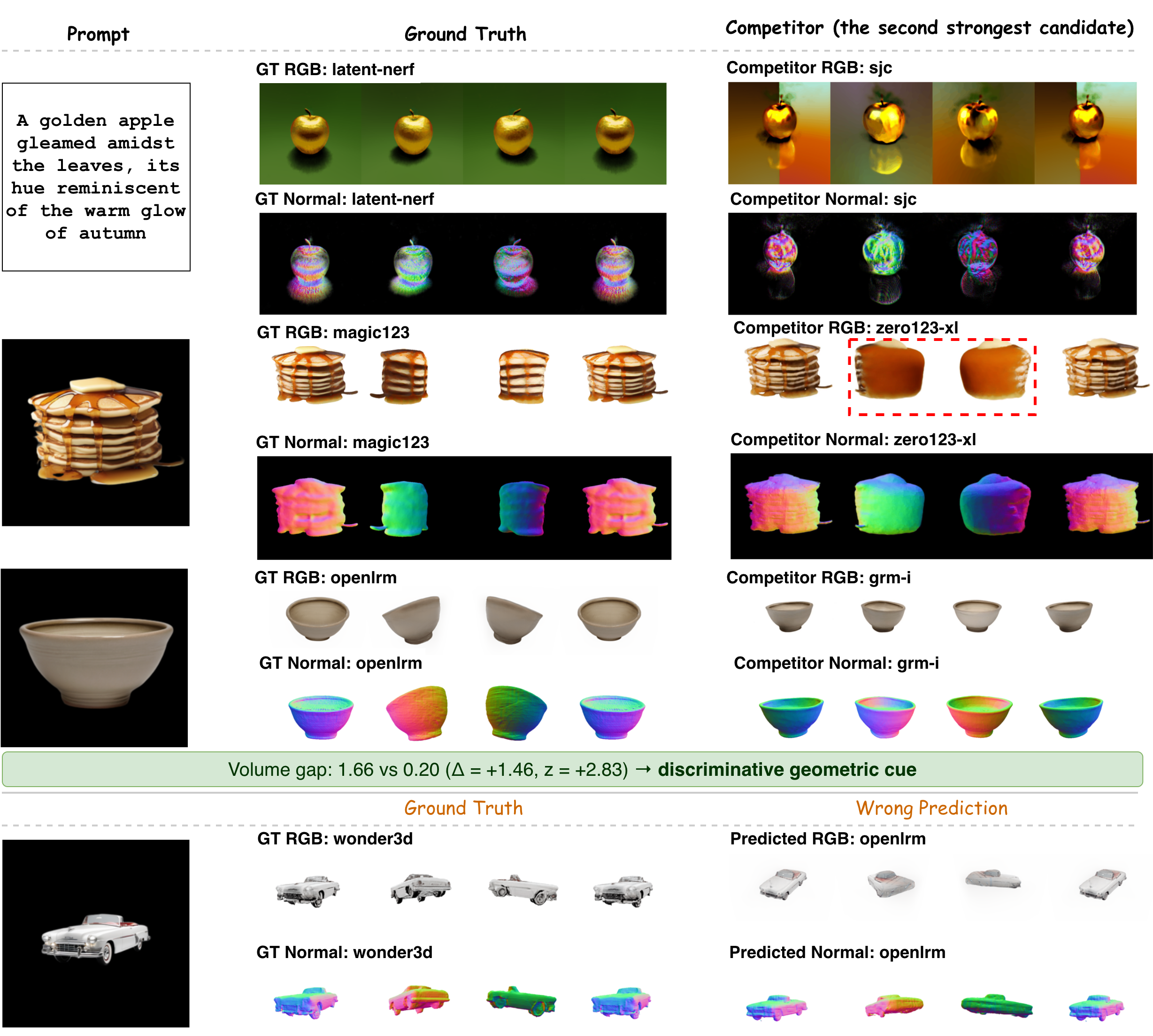}
  \caption{%
    Visualization of attribution fingerprints captured by our framework. 
}
  \label{fig:case-study}
  \vspace{-10pt}
\end{figure}

\subsection{Qualitative Attribution Analysis}\label{sec:experiment-visual-analysis}

Fig.~\ref{fig:case-study} visualizes representative attribution fingerprints by comparing the ground-truth generator against the strongest competing prediction. In the first case, the competing generator already exhibits noticeable rendering differences, providing strong appearance-level attribution signals. In the second case, although several rendered views appear highly similar, the competing model shows severe cross-view inconsistency and unstable geometry in the middle viewpoints, highlighting the importance of explicit cross-view modeling.
The third case demonstrates a more challenging scenario where RGB renderings and normal maps are nearly indistinguishable. Nevertheless, the two generators exhibit substantially different geometry and frequency statistics, providing highly discriminative structural fingerprints beyond appearance alone.
Finally, the failure case shows that attribution becomes considerably harder when generators produce similarly smooth geometry and consistent multi-view structures, suggesting that future attribution systems may require additional fingerprints beyond appearance, geometry, and frequency cues. Additional confusion analysis is in the Appendix.

\section{Conclusion}

We present the first systematic study of passive source attribution for generated 3D assets. We construct a unified benchmark covering $22$ representative 3D generators under standard, few-shot, and realistic deployment settings, and show that modern 3D generators leave stable fingerprints through cross-view inconsistency and structural artifacts in geometry and frequency cues.
Motivated by these findings, we propose a hierarchical multi-view multi-modal Transformer that explicitly models complementary evidence across viewpoints and modalities. Extensive experiments demonstrate strong attribution performance and robustness under realistic deployment scenarios.
Our work establishes a benchmark and methodological foundation for trustworthy 3D content provenance and opens new directions for AI-generated 3D forensics.

\bibliographystyle{unsrt}
\bibliography{references}

\medskip
\newpage

\appendix
\begin{table}[h]
\centering
\caption{Overview of 3D Generation Methods included in the Benchmark}
\label{tab:3d-methods}
\begin{tabular}{lp{10cm}}
\toprule
\textbf{Task} & \textbf{Methods} \\
\midrule
Image-to-3D & Free3D~\cite{Free3D}, Escher-Net~\cite{EscherNet}, Point-E~\cite{Ponit-e}, Triplane-Gaussian~\cite{Triplane}, Shap-E~\cite{Shap-e}, SyncDreamer~\cite{SyncDreamer}, GRM~\cite{GRM}, LGM~\cite{LGM}, Magic123~\cite{Magic123}, Zero123-XL~\cite{Zero-1-to-3}, Stable Zero123~\cite{Zero-1-to-3}, OpenLRM~\cite{OpenLRM}, Wonder3D~\cite{Wonder3D} \\
\midrule
Text-to-3D  & Mvdream~\cite{Mvdream}, Lucid-Dreamer~\cite{Luciddreamer}, Magic3D~\cite{Magic3D}, GRM~\cite{GRM}, Dreamfusion~\cite{dreamfusion}, Latent-NeRF~\cite{Latent-NeRF}, Shap-E~\cite{Shap-e}, SJC~\cite{Score_Jacobian_Chaining}, Point-E~\cite{Ponit-e} \\
\bottomrule
\end{tabular}
\end{table}
\section{Benchmark}\label{sec:appendix-benchmark}

\begin{figure}[t]
  \centering
  \vspace{-4pt}
  \includegraphics[width=0.8\linewidth]{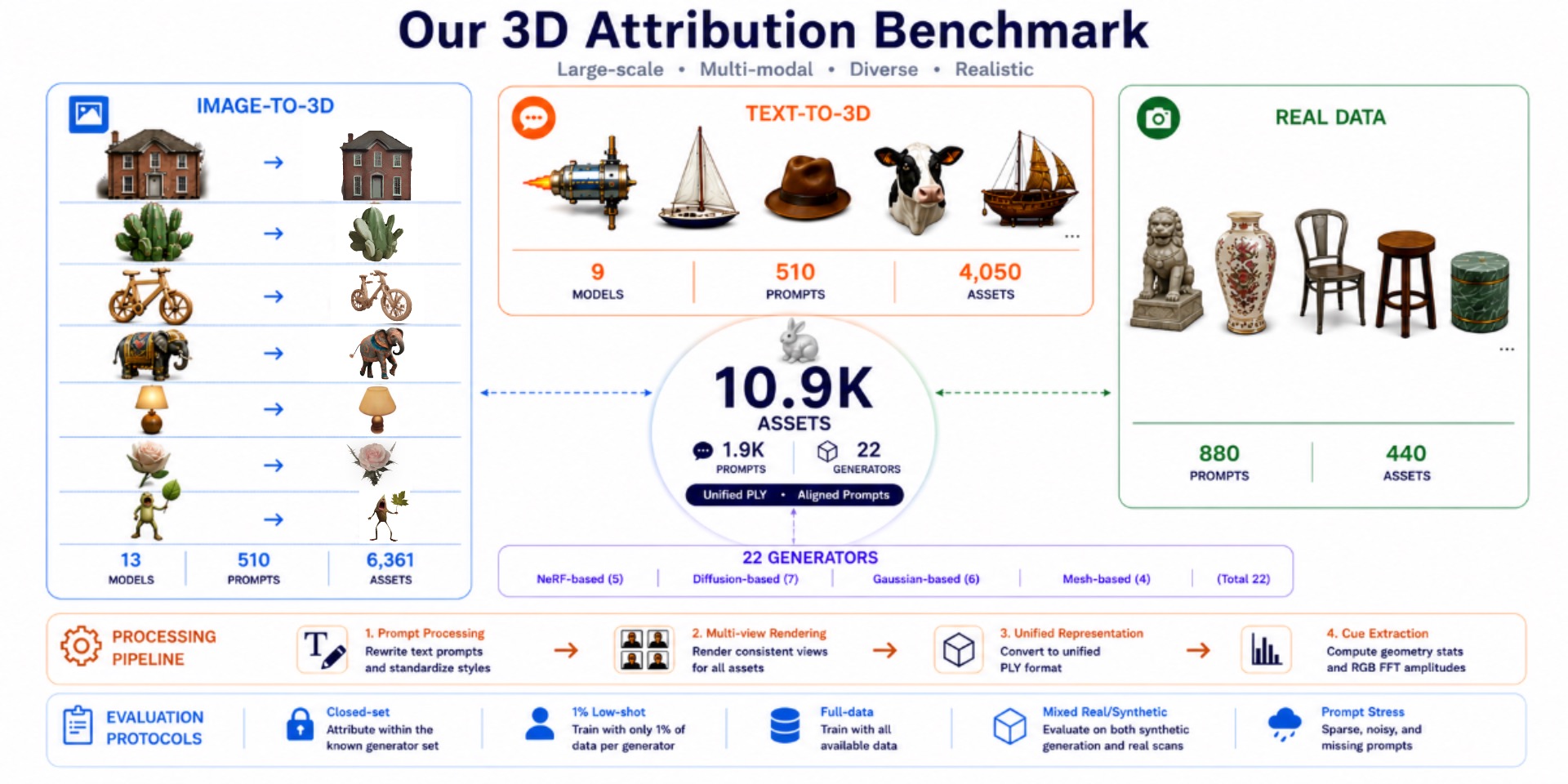}
  \caption{Overview of our attribution benchmark.}
  \label{fig:per-source-comparison}
  \vspace{-16pt}
\end{figure}

\begin{table}[t]
\centering
\scriptsize
\caption{Benchmark summary.}
\label{tab:benchmark_summary}
\begin{tabular}{l l}
\toprule
\textbf{Component} & \textbf{Description} \\
\midrule
Data source & 3DGen-Bench (synthetic) + Cap3D (real) \\
Coverage & 22 generators (text-to-shape \& image-to-shape) \\
Representation & Unified PLY format \\
Derived cues & Geometry statistics + RGB FFT \\
Prompt processing & Rewritten text + rendered image prompts \\
Prompt stress & Sparse, noisy, missing prompts \\
Protocols & Closed-set, 1\% few-shot, mixed real/synthetic, corrupted prompts \\
\midrule
\multicolumn{2}{c}{\textbf{Dataset statistics}} \\
\midrule
Image-to-Shape & 13 models, 510 prompts, 6,361 assets \\
Text-to-Shape & 9 models, 510 prompts, 4,050 assets \\
Real subset & 880 prompts, 440 assets \\
Overall & 22 models, 1,900 prompts, 10,851 assets \\
\bottomrule
\end{tabular}
\end{table}

\noindent\textbf{Overview.}
We construct the first benchmark for 3D asset source attribution, built upon public resources including 3DGen-Bench~\cite{3dgen-bench} and Cap3D~\cite{cap3d}, together with additional components introduced in the following. The benchmark summary is shown in Tab.~\ref{tab:benchmark_summary}.

The synthetic data from 3DGen-Bench~\cite{3dgen-bench} covers outputs from 22 representative generative 3D models, spanning both text-to-3D~\cite{Mvdream,Luciddreamer,Magic3D,GRM,dreamfusion} and image-to-3D~\cite{Free3D,EscherNet,Ponit-e,Triplane,Shap-e} settings, diverse model families, backbone architectures, and underlying 3D representations. Full method list is shown in Tab.~\ref{tab:3d-methods}.
All synthetic assets are provided in a unified PLY format, allowing direct evaluation under a common label space defined by source generator identity.

Beyond raw assets and prompts, we augment the benchmark with attribution-oriented cues, including geometric statistics from the mesh and frequency-domain signals from rendered RGB views~\cite{frank2020leveraging, corvi2023intriguing}.
We further include real scanned assets from Cap3D~\cite{cap3d} dataset, and metadata stress-test variants with incomplete or degraded prompts.
The resulting benchmark supports evaluation under multiple practical regimes, including standard closed-set attribution, few-shot, and realistic deployment protocols, including degraded or missing metadata, and mixed real-world environments.

\noindent\textbf{Harmonization and Processing Pipeline}
To construct a unified attribution benchmark, we further augment inherited public resources with attribution-oriented derived observations and realistic stress-test variants. Specifically, we curate a \textbf{real-data subset} from Cap3D~\cite{cap3d} by selecting assets that are comparable to the synthetic benchmark, rewriting the associated text prompts to align with the benchmark description style, and rendering a canonical RGB view as the corresponding image prompt. To ensure data quality and annotation reliability, each curated sample undergoes cross-validation through manual inspection by at least three independent annotators.
This yields 440 real assets, each paired with one text prompt and one image prompt. We further compute a 102-dim \textbf{geometric descriptor} from mesh geometry, extract Fast Fourier Transform (FFT)-based \textbf{frequency cues} from rendered RGB views~\cite{frank2020leveraging, corvi2023intriguing}, and construct \textbf{metadata stress-test variants} through prompt sparsification, corruption, and masking. 

\begin{table*}[t]
\centering
\footnotesize
\caption{The complete definition of the 102-dimensional geometric feature vector. Abbreviations: \texttt{vtx} vertex count, \texttt{face} face count, \texttt{v/f} vertex-face ratio, \texttt{area} surface area, \texttt{vol} volume approximation, \texttt{ncons} normal consistency, \texttt{cc} connected components, \texttt{wt} watertight, \texttt{mani} manifold, \texttt{wind} winding consistent, \texttt{nmf} nonmanifold-edge fraction, \texttt{bef} boundary-edge fraction, \texttt{sip} self-intersection proxy, and \texttt{sdiam} shape diameter.}
\setlength{\tabcolsep}{6pt}
\renewcommand{\arraystretch}{1.10}
\begin{tabular}{l p{6cm} c p{4cm}}
\toprule
\textbf{Group} & \textbf{Items} & \textbf{Dim} & \textbf{Meaning} \\
\hline
\texttt{count} & \texttt{vtx}, \texttt{face}, \texttt{v/f} & 3 & Mesh size and density. \\
\texttt{bbox} & \texttt{bx}, \texttt{by}, \texttt{bz}, \texttt{bx/by}, \texttt{by/bz}, \texttt{bx/bz}, \texttt{bmax/bmin} & 7 & Global scale and aspect ratio. \\
\texttt{topo+shape} & \texttt{area}, \texttt{vol}, \texttt{ncons}, \texttt{cc}, \texttt{wt}, \texttt{mani}, \texttt{wind}, \texttt{euler}, \texttt{nmf}, \texttt{bef}, \texttt{sip}, \texttt{sdiam} & 12 & Bulk, smoothness, connectivity, topology, and extent. \\
\texttt{edge-hist} & \texttt{edge\_hist[16]} & 16 & Edge-length distribution. \\
\texttt{face-hist} & \texttt{face\_hist[16]} & 16 & Face-area distribution. \\
\texttt{curv-hist} & \texttt{curv\_hist[16]} & 16 & Curvature distribution. \\
\texttt{spec} & \texttt{eig[16]} & 16 & Laplacian spectrum. \\
\texttt{dist-hist} & \texttt{dist\_hist[16]} & 16 & Surface-distance distribution. \\
\hline
\textbf{Total} & 22 scalar + 64 histogram + 16 spectral features & \textbf{102} & $22 + 64 + 16 = 102$. \\
\bottomrule
\end{tabular}
\label{tab:geom-102d-compact-star}
\end{table*}

\noindent\textbf{Geometric Representation.}
The 102-dim geometric vector consists of 22 scalar statistics, four 16-bin histograms, and 16 Laplacian eigenvalue features, giving $22 + 4\times 16 + 16 = 102$. Full definition is listed in Tab.~\ref{tab:geom-102d-compact-star}.

\begin{table*}[t]
\centering
\footnotesize
\caption{The complete definition of the RGB FFT feature.}
\setlength{\tabcolsep}{6pt}
\renewcommand{\arraystretch}{1.10}
\begin{tabular}{l l c l}
\toprule
\textbf{Group} & \textbf{Items} & \textbf{Dim} & \textbf{Meaning} \\
\hline
\texttt{gray} & mean over RGB channels & 1 map & Convert RGB image to a single grayscale image before FFT. \\
\texttt{fft} & 2D FFT magnitude & 1 map & Capture frequency energy over spatial frequencies. \\
\texttt{log-amp} & \texttt{log(1 + |FFT|)} & 1 map & Compress dynamic range and stabilize large peaks. \\
\texttt{shift} & centered spectrum & 1 map & Move low frequency to the center for spatially aligned pooling. \\
\texttt{pool16} & \texttt{avgpool}($16 \times 16$) & 256 & Downsample the FFT amplitude map into a compact frequency grid. \\
\hline
\textbf{Per-view total} & \texttt{fft\_map[16,16]} & \textbf{256} & Raw RGB-FFT feature for one view. \\
\hline
\textbf{Total} & 4 views $\times$ 256 dims/view & \textbf{1024} & Raw RGB-FFT input across 4 views before token projection. \\
\bottomrule
\end{tabular}
\label{tab:rgb-fft-256d-compact-star}
\end{table*}

\noindent\textbf{Domain frequency data.}
For each view, the RGB image is averaged to grayscale, transformed by a 2D FFT, converted to a centered log-amplitude spectrum, and adaptively average-pooled to a $16 \times 16$ map. This gives a 256-dimensional raw frequency descriptor per view. In the default 4-view PH6 setup, the raw RGB-FFT branch therefore contributes $4 \times 256 = 1024$ values before linear token projection. Full definition is listed in Tab.~\ref{tab:rgb-fft-256d-compact-star}.

\section{Method}\label{sec:appendix-method}
\noindent\textbf{Implementation details.}
All models are trained using AdamW~\cite{adamw} optimizer with cosine learning-rate decay. The initial learning rate is set to $1\times10^{-4}$ with weight decay $1\times10^{-2}$. Models are trained for 100 epochs using a batch size of 32. During training, metadata inputs may be randomly dropped with a fixed probability to improve robustness under missing-prompt deployment scenarios. Standard data augmentation including random horizontal flipping and color jittering is applied to RGB renderings.All experiments are implemented in PyTorch and trained on NVIDIA H100 GPUs.

\section{Experiment}\label{sec:appendix-experiment}

\begin{table}[ht]
  \centering
  \footnotesize
  \setlength{\tabcolsep}{8pt}
  \caption{Per-source performance analysis across different generative models. \colorbox{easygreen}{Green} and \colorbox{hardred}{red} cells indicate highly reliable and challenging attribution cases, respectively.}
  \label{tab:appendix-per-source}
  \begin{tabular}{ll cccc cccc}
    \toprule
    & & \multicolumn{4}{c}{1\% Data} & \multicolumn{4}{c}{Full Data} \\
    \cmidrule(r){3-6} \cmidrule(l){7-10}
    Model & Task & Prec. & Rec. & F1 & Acc. & Prec. & Rec. & F1 & Acc. \\
    \midrule
    L-Dreamer~\cite{Luciddreamer} & Text & \cellcolor{green!12}{100.0} & \cellcolor{green!12}{97.47} & \cellcolor{green!12}{98.72} & \cellcolor{green!12}{100.0} & \cellcolor{green!12}{100.0} & \cellcolor{green!12}{100.0} & \cellcolor{green!12}{100.0} & \cellcolor{green!12}{100.0} \\
    Point-E-T~\cite{Ponit-e}    & Text & 88.46 & 58.47 & 70.41 & 88.46 & \cellcolor{green!12}{100.0} & 88.46 & \cellcolor{green!12}{93.88} & 88.46 \\
    Latent-NeRF~\cite{Latent-NeRF}    & Text & 86.42 & 72.92 & 79.10 & 86.42 & \cellcolor{green!12}{91.95} & \cellcolor{green!12}{98.77} & \cellcolor{green!12}{95.24} & \cellcolor{green!12}{98.77} \\
    SJC~\cite{Score_Jacobian_Chaining}       & Text & 82.00 & 67.21 & 73.87 & 82.00 & \cellcolor{green!12}{100.0} & \cellcolor{green!12}{100.0} & \cellcolor{green!12}{100.0} & \cellcolor{green!12}{100.0} \\
    GRM-T~\cite{GRM}     & Text & 75.34 & \cellcolor{green!12}{94.83} & 83.97 & 75.34 & \cellcolor{green!12}{100.0} & \cellcolor{green!12}{100.0} & \cellcolor{green!12}{100.0} & \cellcolor{green!12}{100.0} \\
    Magic3D~\cite{Magic3D}   & Text & 78.95 & 80.00 & 79.47 & 78.95 & \cellcolor{green!12}{98.68} & \cellcolor{green!12}{98.68} & \cellcolor{green!12}{98.68} & \cellcolor{green!12}{98.68} \\
    MVDream~\cite{Mvdream}   & Text & 78.67 & \cellcolor{green!12}{93.65} & 85.51 & 78.67 & \cellcolor{green!12}{98.59} & \cellcolor{green!12}{93.33} & \cellcolor{green!12}{95.89} & \cellcolor{green!12}{93.33} \\
    Shap-E-T~\cite{Shap-e}    & Text & \cellcolor{red!10}{26.32} & 86.96 & 40.40 & \cellcolor{red!10}{26.32} & 89.41 & \cellcolor{green!12}{100.0} & \cellcolor{green!12}{94.41} & \cellcolor{green!12}{100.0} \\
    DreamFusion~\cite{dreamfusion}   & Text & \cellcolor{red!10}{6.41} & 83.33 & \cellcolor{red!10}{11.90} & \cellcolor{red!10}{6.41} & \cellcolor{green!12}{97.40} & \cellcolor{green!12}{96.15} & \cellcolor{green!12}{96.77} & \cellcolor{green!12}{96.15} \\
    \midrule
    Overall & Text & 70.79 & 69.17 & 69.97 & 68.67 & \cellcolor{green!12}{97.48} & \cellcolor{green!12}{97.27} & \cellcolor{green!12}{97.28} & \cellcolor{green!12}{97.14} \\
    \midrule
    Triplane-Gau.~\cite{Triplane}  & Image & \cellcolor{green!12}{98.77} & \cellcolor{green!12}{94.12} & \cellcolor{green!12}{96.39} & \cellcolor{green!12}{98.77} & \cellcolor{green!12}{100.0} & \cellcolor{green!12}{100.0} & \cellcolor{green!12}{100.0} & \cellcolor{green!12}{100.0} \\
    OpenLRM~\cite{OpenLRM}   & Image & \cellcolor{green!12}{96.43} & 79.41 & 87.10 & \cellcolor{green!12}{96.43} & \cellcolor{green!12}{100.0} & \cellcolor{green!12}{100.0} & \cellcolor{green!12}{100.0} & \cellcolor{green!12}{100.0} \\
    Free3D~\cite{Free3D}    & Image & \cellcolor{green!12}{96.25} & 83.70 & 89.53 & \cellcolor{green!12}{96.25} & \cellcolor{green!12}{96.30} & \cellcolor{green!12}{97.50} & \cellcolor{green!12}{96.89} & \cellcolor{green!12}{97.50} \\
    GRM-I~\cite{GRM}     & Image & \cellcolor{green!12}{95.18} & 73.15 & 82.72 & \cellcolor{green!12}{95.18} & \cellcolor{green!12}{100.0} & \cellcolor{green!12}{100.0} & \cellcolor{green!12}{100.0} & \cellcolor{green!12}{100.0} \\
    LGM~\cite{LGM}       & Image & \cellcolor{green!12}{91.46} & \cellcolor{green!12}{96.15} & \cellcolor{green!12}{93.75} & \cellcolor{green!12}{91.46} & \cellcolor{green!12}{100.0} & \cellcolor{green!12}{100.0} & \cellcolor{green!12}{100.0} & \cellcolor{green!12}{100.0} \\
    Magic123~\cite{Magic123}  & Image & \cellcolor{green!12}{91.46} & 64.10 & 75.38 & \cellcolor{green!12}{91.46} & \cellcolor{green!12}{98.80} & \cellcolor{green!12}{100.0} & \cellcolor{green!12}{99.39} & \cellcolor{green!12}{100.0} \\
    Point-E-I~\cite{Ponit-e}    & Image & 89.02 & 82.02 & 85.38 & 89.02 & \cellcolor{green!12}{97.56} & \cellcolor{green!12}{97.56} & \cellcolor{green!12}{97.56} & \cellcolor{green!12}{97.56} \\
    Escher-Net~\cite{EscherNet}    & Image & 89.02 & 66.97 & 76.44 & 89.02 & \cellcolor{green!12}{96.43} & \cellcolor{green!12}{98.78} & \cellcolor{green!12}{97.59} & \cellcolor{green!12}{98.78} \\
    Shap-E-I~\cite{Shap-e}    & Image & 81.48 & 67.35 & 73.74 & 81.48 & \cellcolor{green!12}{95.18} & \cellcolor{green!12}{97.53} & \cellcolor{green!12}{96.34} & \cellcolor{green!12}{97.53} \\
    Zero123-XL~\cite{Zero-1-to-3}   & Image & 76.25 & 64.21 & 69.71 & 76.25 & 89.41 & \cellcolor{green!12}{95.00} & \cellcolor{green!12}{92.12} & \cellcolor{green!12}{95.00} \\
    SyncDreamer~\cite{SyncDreamer}   & Image & 75.00 & 82.89 & 78.75 & 75.00 & \cellcolor{green!12}{100.0} & \cellcolor{green!12}{94.05} & \cellcolor{green!12}{96.93} & \cellcolor{green!12}{94.05} \\
    Wonder3D~\cite{Wonder3D}  & Image & 45.68 & 84.09 & 59.20 & 45.68 & \cellcolor{green!12}{96.30} & \cellcolor{green!12}{96.30} & \cellcolor{green!12}{96.30} & \cellcolor{green!12}{96.30} \\
    Stable-Zero123~\cite{Zero-1-to-3}    & Image & 45.00 & 66.67 & 53.73 & 45.00 & \cellcolor{green!12}{95.89} & 87.50 & \cellcolor{green!12}{91.50} & 87.50 \\
    \midrule
    Overall & Image & 83.27 & 82.39 & 82.16 & 82.49 & \cellcolor{green!12}{97.37} & \cellcolor{green!12}{97.25} & \cellcolor{green!12}{97.28} & \cellcolor{green!12}{97.27} \\
    \midrule
    Average & -- & 79.08 & 76.98 & 71.53 & 77.17 & \cellcolor{green!12}{93.13} & \cellcolor{green!12}{93.03} & \cellcolor{green!12}{93.02} & \cellcolor{green!12}{97.22} \\
    \bottomrule
  \end{tabular}
\end{table}

\noindent\textbf{Per-source Attribution analysis.}
Tab.~\ref{tab:appendix-per-source} provides full per-generator attribution results under both 1\% and full-data settings. Overall, our model achieves strong and consistent attribution performance across diverse generator families, while also revealing substantial differences in attribution difficulty across models.

Under the 1\% setting, image-to-3D generators are generally more distinguishable than text-to-3D generators. Several image-conditioned models, including Triplane-Gaussian~\cite{Triplane}, OpenLRM~\cite{OpenLRM}, Free3D~\cite{Free3D}, GRM-I~\cite{GRM}, and LGM~\cite{LGM}, already exceed 90\% accuracy with extremely limited supervision, suggesting that these generators leave highly stable structural and cross-view fingerprints. In contrast, text-to-3D generators exhibit substantially larger performance variation, indicating that open-ended geometry generation introduces more diverse and unstable attribution patterns.

The most challenging cases arise from generators with severe cross-view inconsistency, coarse geometry, or unstable topology. In particular, DreamFusion~\cite{dreamfusion} and Shap-E-T~\cite{Shap-e} remain difficult under low-data supervision, achieving only 6.41\% and 26.32\% accuracy, respectively. These generators frequently produce fragmented structures, unstable hidden-surface completion, and irregular geometry across views, making attribution substantially harder for appearance-dominant baselines. Similar trends are observed for Wonder3D~\cite{Wonder3D} and Stable-Zero123~\cite{Zero-1-to-3} under image-conditioned settings.

Importantly, the performance gap between the 1\% and full-data settings demonstrates that our model effectively captures stable generator-specific fingerprints once sufficient supervision becomes available. Many generators improve from moderately separable to nearly perfect attribution under full-data training, with multiple models reaching or approaching 100\% accuracy. This trend is particularly pronounced for structurally challenging generators such as DreamFusion~\cite{dreamfusion}, Shap-E~\cite{Shap-e}, and Wonder3D~\cite{Wonder3D}, suggesting that the proposed hierarchical multi-view modeling progressively learns robust attribution cues beyond superficial rendering appearance.

We also observe that confusion patterns are highly structured rather than random. Generators belonging to similar architectural families or sharing related geometric priors tend to remain more difficult to separate, while generators with distinctive cross-view artifacts or topology statistics are attributed much more reliably. Additional confusion analysis are provided in the following parts.

\begin{wrapfigure}{r}{0.35\textwidth}
  \centering
  \vspace{-8pt}
  \includegraphics[width=0.33\textwidth]{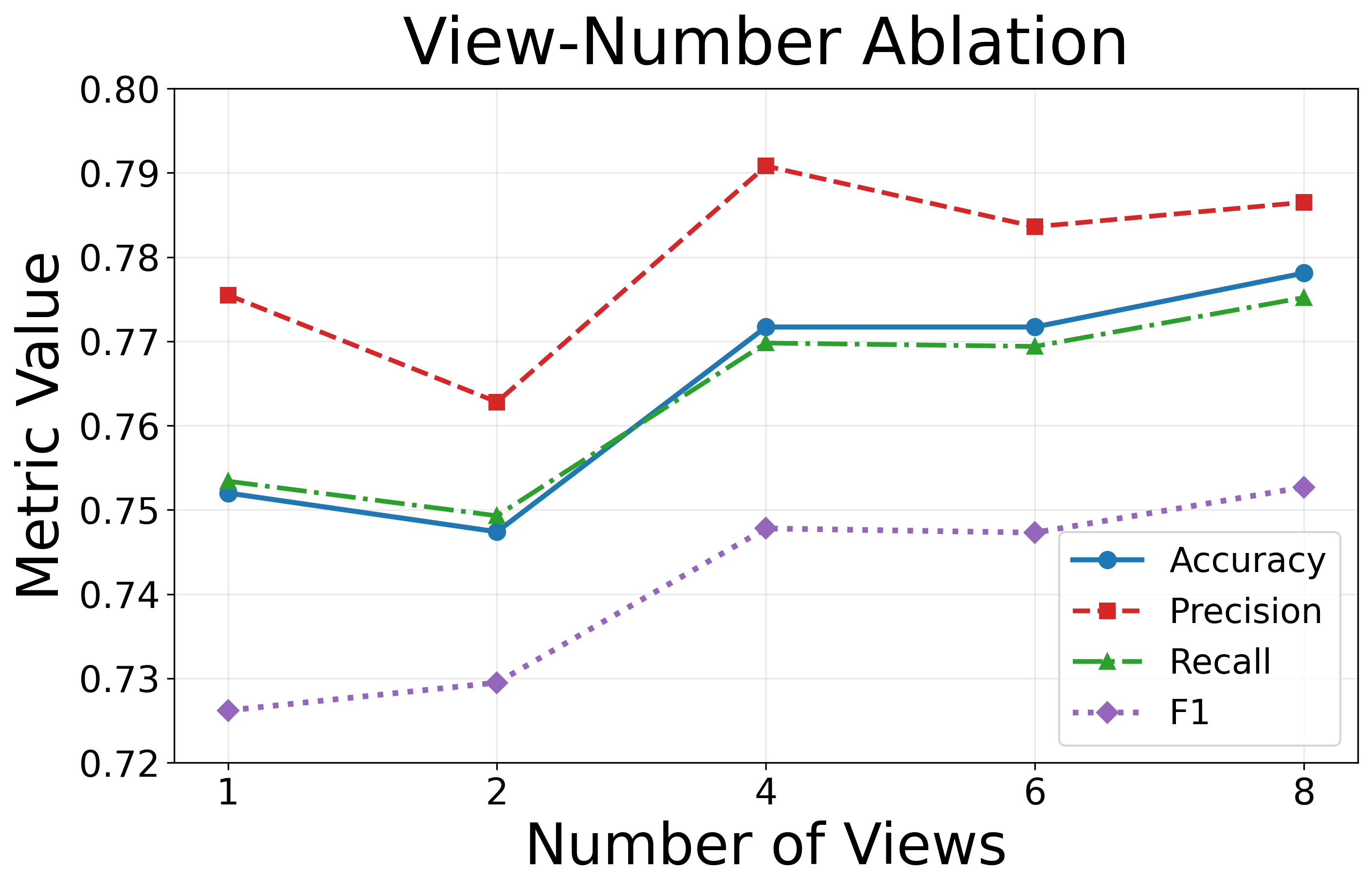}
  \caption{View number ablation.}
  \label{fig:view-num-ablation}
  \vspace{-10pt}
\end{wrapfigure}

\noindent\textbf{View number ablation study.}
We further study the effect of the number of rendered views in Fig.~\ref{fig:view-num-ablation}. Increasing the number of views improves attribution performance from one to four views, after which gains largely saturate. In particular, F1 increases from 72.6 to 74.8 when expanding from one to four views, but only marginally improves thereafter, indicating that four views already capture most useful attribution evidence. More importantly, this gain remains smaller than the improvement brought by explicit multi-view interaction in Tab.~\ref{tab:ablation-study} (+4.88 F1), suggesting that how multiple views are modeled is more important than simply increasing their number. These results validate our default choice of four views and motivate structured cross-view reasoning over brute-force view expansion.

\begin{table}[t]
\centering
\small
\caption{Robust attribution under degraded and missing metadata. ``Empty'' denotes missing prompts only during inference, while ``Empty*'' removes prompts during both training and inference.}
\label{tab:appendix-prompt-dependency}
\begin{tabular}{clcccc}
\toprule
Task & Setting & Prec. & Rec. & F1 & Acc. \\
\midrule

\multirow{5}{*}{Sparse Text Prompt}
& Full        & 85.06 & 69.17 & 69.97 & 68.67 \\
& Sparse (4 words) & 85.35 & 68.47 & 70.25 & 67.92 \\
& Sparse (1 word) & 85.09 & 64.16 & 67.27 & 63.55 \\
& Empty       & 87.29 & 63.74 & 67.08 & 63.25 \\
& Empty*      & 84.14 & 61.50 & 66.24 & 60.69 \\

\midrule

\multirow{8}{*}{
\begin{tabular}{c}
Noisy Image Prompt \\
{\small (Gaussian noise std. $\sigma$)}
\end{tabular}
}
& Clean        & 84.15 & 82.39 & 82.16 & 82.49 \\
& $\sigma=8$   & 84.39 & 82.57 & 82.42 & 82.67 \\
& $\sigma=16$  & 84.57 & 82.76 & 82.74 & 82.86 \\
& $\sigma=32$  & 85.15 & 82.38 & 82.96 & 82.49 \\
& $\sigma=48$  & 84.88 & 81.62 & 82.40 & 81.73 \\
& $\sigma=64$  & 85.25 & 81.62 & 82.53 & 81.73 \\
& $\sigma=96$  & 85.53 & 81.81 & 82.71 & 81.92 \\
& Empty*       & 84.27 & 78.62 & 80.10 & 78.72 \\

\midrule

\multirow{10}{*}{Masked Image Prompt ($r$)}
& $r=5\%$  & 49.94 & 48.85 & 48.75 & 82.77 \\
& $r=10\%$ & 49.61 & 48.51 & 48.37 & 82.20 \\
& $r=20\%$ & 50.43 & 48.75 & 49.02 & 82.58 \\
& $r=30\%$ & 50.17 & 48.41 & 48.60 & 82.02 \\
& $r=40\%$ & 50.09 & 47.74 & 48.26 & 80.89 \\
& $r=50\%$ & 50.67 & 47.14 & 48.24 & 79.85 \\
& $r=60\%$ & 50.01 & 45.91 & 47.29 & 77.78 \\
& $r=70\%$ & 50.30 & 44.57 & 46.52 & 75.52 \\
& $r=80\%$ & 50.14 & 44.07 & 46.24 & 74.67 \\
& $r=90\%$ & 50.72 & 43.69 & 46.19 & 74.01 \\

\bottomrule
\end{tabular}
\end{table}

\noindent\textbf{Robust attribution under realistic deployment settings.}
Tab.~\ref{tab:appendix-prompt-dependency} evaluates our model under degraded and missing metadata conditions, including sparse text prompts, noisy image prompts, and masked image prompts. Overall, the model remains highly robust even when metadata quality is severely degraded, suggesting that attribution primarily relies on stable structural and cross-view fingerprints rather than prompt-specific shortcuts.

For sparse text prompts, performance degrades gradually as textual information becomes increasingly limited. Reducing prompts from full descriptions to only four words causes almost no performance drop (68.67\% $\rightarrow$ 67.92\% accuracy), while even single-word prompts still preserve meaningful attribution capability (63.55\% accuracy). Notably, when prompts are entirely removed during inference (Empty), the model still achieves 63.25\% accuracy, and remains robust even when prompts are unavailable during both training and inference (Empty*), reaching 60.69\% accuracy. These results suggest that our model learns generator-specific fingerprints beyond metadata supervision alone.

For image-conditioned attribution, our model is remarkably insensitive to prompt corruption. Across a wide range of Gaussian noise levels ($\sigma=8$ to $\sigma=96$), attribution performance remains nearly unchanged, with accuracy consistently around 82\%. Even under severe corruption ($\sigma=96$), the model only drops from 82.49\% to 81.92\% accuracy, indicating that generator-specific structural and frequency cues dominate over prompt fidelity.

We further evaluate robustness under progressively masked image prompts. Despite substantial prompt removal, our model maintains strong attribution capability across all masking ratios. Accuracy remains above 80\% up to $r=40\%$ masking and still achieves 74.01\% accuracy even when 90\% of the image prompt is removed. The relatively smooth degradation trend suggests that the model does not rely on isolated local appearance patterns, but instead captures more globally stable attribution fingerprints distributed across views and structural observations.

Overall, these results demonstrate that our model generalizes reliably under realistic deployment conditions where metadata may be sparse, corrupted, partially missing, or entirely unavailable.

\begin{figure}
  \centering
  \vspace{-8pt}
  \includegraphics[width=0.6\linewidth]{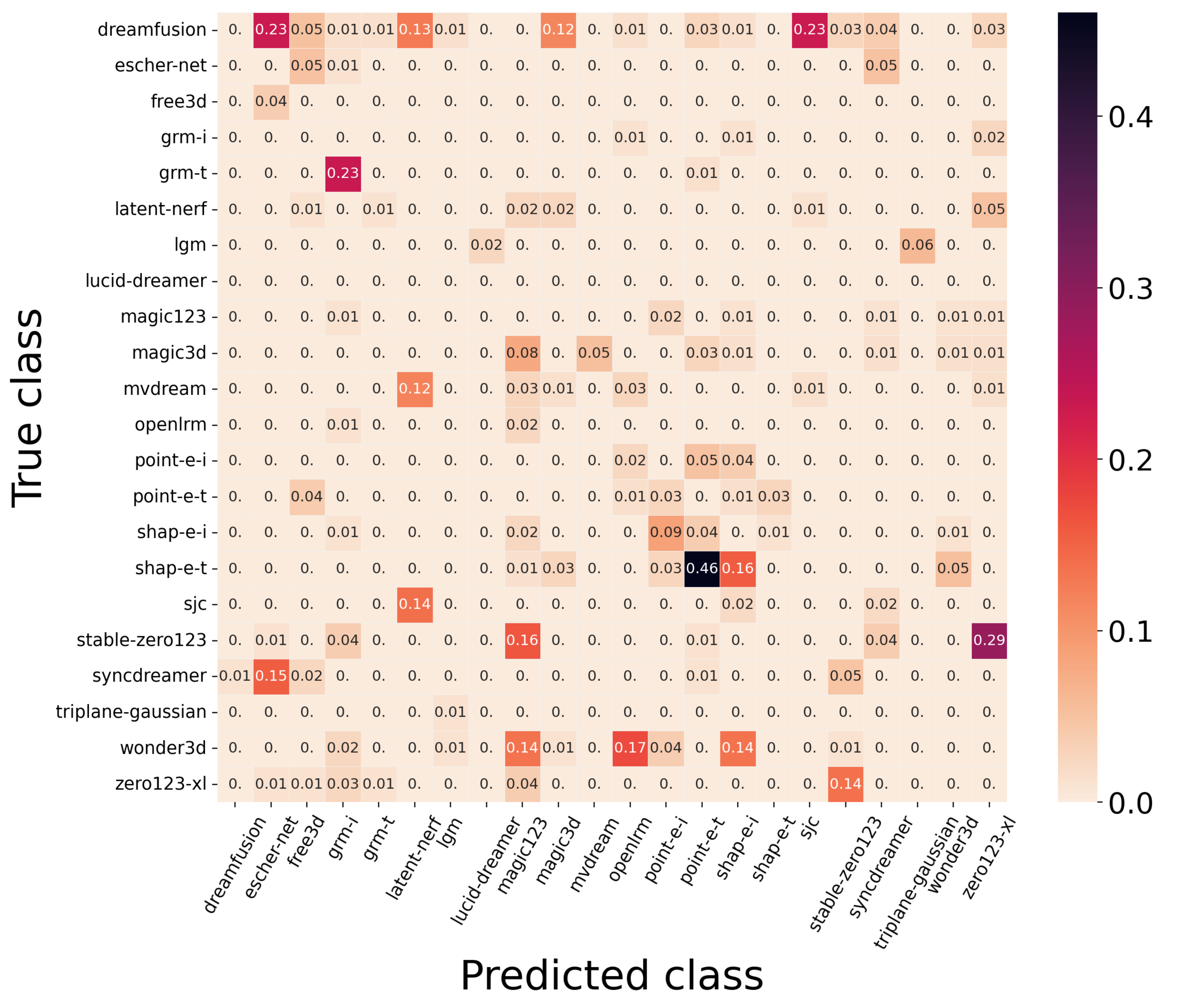}
  \caption{Row-normalized confusion matrix heatmap. Each value denotes the proportion of samples from the ground-truth generator (row) that are misattributed to the predicted generator (column). Higher values indicate stronger inter-generator confusion patterns.}
  \label{fig:confusion-matrix}
  \vspace{-10pt}
\end{figure}

\noindent\textbf{Error analysis.}
Fig.~\ref{fig:confusion-matrix} presents the row-normalized confusion matrix across generators. Overall, attribution errors are highly structured rather than random, suggesting that our model captures meaningful generator-level relationships in the learned attribution space.

The most significant confusions occur between generators sharing similar rendering styles, geometric priors, or architectural families. For example, Shap-E-T~\cite{Shap-e} is frequently confused with Point-E-T ($0.46$), likely because both generators often produce coarse and incomplete geometry with unstable topology. Similarly, Stable-Zero123~\cite{Zero-1-to-3} exhibits strong confusion with Zero123-XL~\cite{Zero-1-to-3} ($0.29$), reflecting their closely related model families and highly similar multi-view consistency patterns.

We also observe confusion between DreamFusion~\cite{dreamfusion} and EscherNet~\cite{EscherNet} ($0.23$), as well as GRM-T~\cite{GRM} and GRM-I~\cite{GRM} ($0.23$), suggesting that generators with similar rendering smoothness or geometry regularization strategies may leave partially overlapping attribution fingerprints. Wonder3D~\cite{Wonder3D} further exhibits distributed confusion across multiple image-conditioned generators, indicating that highly regularized geometry and smooth surface priors can reduce attribution separability.

Importantly, most generator pairs exhibit near-zero off-diagonal confusion, demonstrating that the proposed hierarchical multi-view modeling learns highly discriminative structural and cross-view fingerprints for the majority of generators. The remaining failure cases primarily arise from structurally similar generators rather than random prediction failures, highlighting a promising direction for future attribution research.

\end{document}